\documentclass{article}
\usepackage{spconf,amsmath,graphicx}

\usepackage[compress]{cite}
\usepackage{enumitem}
\setlist{nosep, leftmargin=14pt}
\usepackage{subfigure}
\usepackage{mwe} 
\usepackage[T1]{fontenc}
\usepackage{multirow}
\usepackage{amsmath,amsfonts,amssymb}
\usepackage{makecell}
\usepackage[colorlinks=true, allcolors=blue]{hyperref}
\usepackage{diagbox}

\title{Metal-conscious Embedding for CBCT Projection Inpainting}
\name{
\begin{tabular}{c}
Fuxin~Fan$^{1*}$,\thanks{$^{*}$ Email: fuxin.fan@fau.de} Yangkong~Wang$^{1}$, Ludwig~Ritschl$^{2}$, Ramyar~Biniazan$^{2}$, Marcel~Beister$^{2}$, \\Bj\"orn Kreher$^{2}$, Yixing~Huang$^{3}$, Steffen Kappler$^{2}$, and Andreas~Maier$^{1}$
\end{tabular}
}
\address{$^{1}$ Pattern Recognition Lab,
Friedrich-Alexander-University Erlangen-Nuremberg, 
Germany\\
$^{2}$ Siemens Healthineers AG,
Forchheim,
Germany\\
$^{3}$ Department of Radiation Oncology,
Friedrich-Alexander-University Erlangen-Nuremberg, 
Germany
}
\begin{document}
%
\maketitle
\begin{abstract}
The existence of metallic implants in projection images for cone-beam computed tomography (CBCT) introduces undesired artifacts which degrade the quality of reconstructed images. In order to reduce metal artifacts, projection inpainting is an essential step in many metal artifact reduction algorithms. In this work, a hybrid network combining the shift window (Swin) vision transformer (ViT) and a convolutional neural network is proposed as a baseline network for the inpainting task. To incorporate metal information for the Swin ViT-based encoder, metal-conscious self-embedding and neighborhood-embedding methods are investigated. Both methods have improved the performance of the baseline network. Furthermore, by choosing appropriate window size, the model with neighborhood-embedding could achieve the lowest mean absolute error of 0.079 in metal regions and the highest peak signal-to-noise ratio of 42.346 in CBCT projections. At the end, the efficiency of metal-conscious embedding on both simulated and real cadaver CBCT data has been demonstrated, where the inpainting capability of the baseline network has been enhanced.
\end{abstract}
\section{Introduction}
\label{sec:intro}
Metallic tools like screws and plates are used in leg surgery for the purpose of bone fixation \cite{luo2013stress}. In the process of metal placement, cone-beam computed tomography (CBCT) devices are often used to guarantee the accuracy \cite{kernen2016accuracy}. However, the accompanying metal artifacts will degrade the image quality because of the high attenuation of such metallic tools \cite{katsura2018current}. Many metal artifact reduction (MAR) algorithms have been proposed, which can be defined into three categories depending on the domain the algorithms work with. Sinogram or projection domain based algorithms tend to complete the missing part of sinograms or projections before image reconstruction \cite{meyer2010normalized, meyer2012frequency, ghani2018deep, liao2019generative, gottschalk2022dl}. Algorithms based on reconstructed volume use deep learning methods to reduce the artifacts in CT images directly \cite{huang2018metal, wang2021dicdnet}, but they are not suitable for CBCT images with extra cone beam artifacts and truncation artifacts. Dual domain based algorithms tackle the problem in both sinograms and volume slices \cite{zhang2018convolutional, wang2021dual}, in which higher computational consumption is needed when applied to CBCT. Therefore, projection inpainting appears a better solution to MAR of CBCT, which is the focus of this work.

Despite the simplicity of the interpolation methods, such algorithms fail to generate structure related textures in missing area when they are used for the inpainting task. Convolutional neural networks (CNNs) have been reported to have better estimation of the correlation between the background and missing area, therefore they were successful in projection inpainting \cite{liao2019generative, gottschalk2022dl,agrawal2021metal}.
However, kernel-based convolutional layers only focus on the relatively local correlations. In this case, the inpainting of missing areas highly depends on their surroundings but not the whole region. Recently vision transformer (ViT) based networks become popular because of the attention mechanism \cite{dosovitskiy2021an}. The usage of the shift window (Swin) which defines the region for attention calculation \cite{liu2022swin} further improves ViT in dense inference applications. Considering the patch-wise modeling properties and the long-range dependencies of ViT, global correlations can be achieved. The works in \cite{Li_2022_CVPR, Dong_2022_CVPR} have shown the efficacy of ViT-based networks in image inpainting task.
However, due to the high resolution of CBCT projections, ViT training will be problematic and the results will not be adequate. 

\begin{figure*}[tb!]
   \begin{center}
   \begin{tabular}{c} 
\includegraphics[width=11cm]{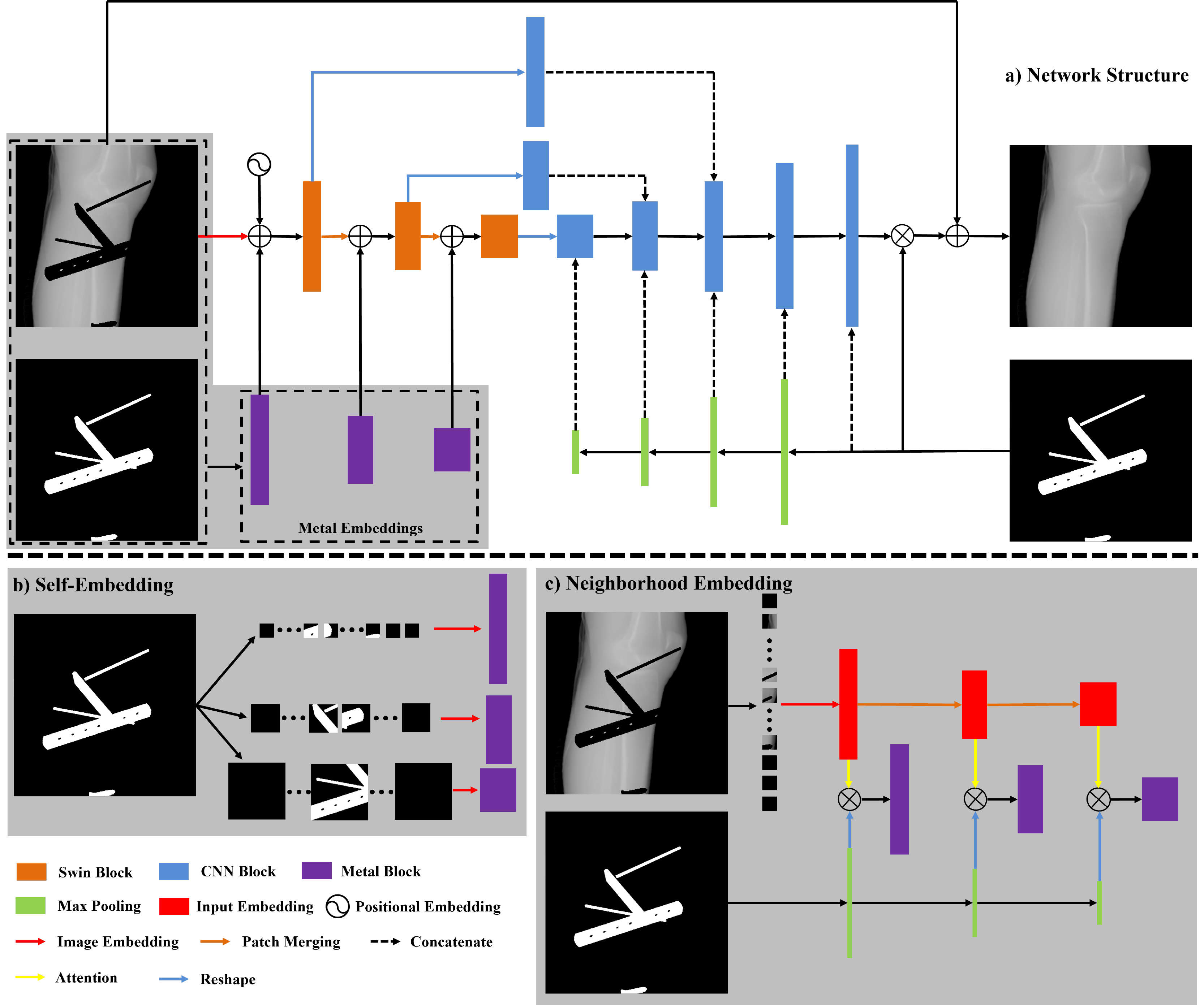}
   \end{tabular}
   \end{center}
   \caption[figure1] 
   { \label{fig:figure1} 
The hybrid Swin-CNN network structure (a) with the metal-conscious embedding layers (purple blocks). The lower part shows the detailed structures of the self-embedding (b) and the neighborhood embedding (c).}
\end{figure*}
Motivated by what is stated above, the Swin ViT is preferred in this work for the correlation calculation.
In this case, we propose a hybrid baseline network with Swin ViT and CNN for CBCT projection inpainting. To complement the baseline network with additional metal information, metal-conscious self-embedding (SE) and neighborhood embedding (NE) methods are investigated. We test our models on both simulation and real cadaver data sets, and the results are compared among different models.

\section{Materials and method}
\label{sec:method}
\subsection{Hybrid network with metal-conscious embedding}
To benefit the network from both local and global correlation calculation, the hybrid Swin-CNN network is proposed and displayed in Fig.~\ref{fig:figure1}. The baseline method without metal-conscious embedding has cascaded Swin blocks as the encoder, the CNN blocks as the decoder and the skip connections in between. The input image is split to patches with the size of 4\,$\times$\,4 and each patch is embedded to 128 channels. Together with absolute positional embedding, the sequence is fed into Swin blocks with patch merging. The number of layers in the first Swin block is 2 and it increases to 4 and 6 after two times of patch merging. The channel size and the number of heads are also doubled after patch merging, with the initial head number of 8. The window size for all Swin blocks are 8 and the even layers in Swin blocks have the shift window size of 4. The output of Swin blocks are reshaped and then fed into convolutional layers to guarantee the right concatenation with CNN-based decoder. Besides, the metal mask is added to the corresponding CNN blocks after maxpooling, which is similar to the structure of the mask pyramid network (MPN) \cite{liao2019generative}. The output has ReLU as active function and the loss function is the combination of mean absolute error (MAE) and adversarial loss.
\begin{figure}[htb]
   \begin{center}
   \begin{tabular}{c} 
   \includegraphics[width=5cm]{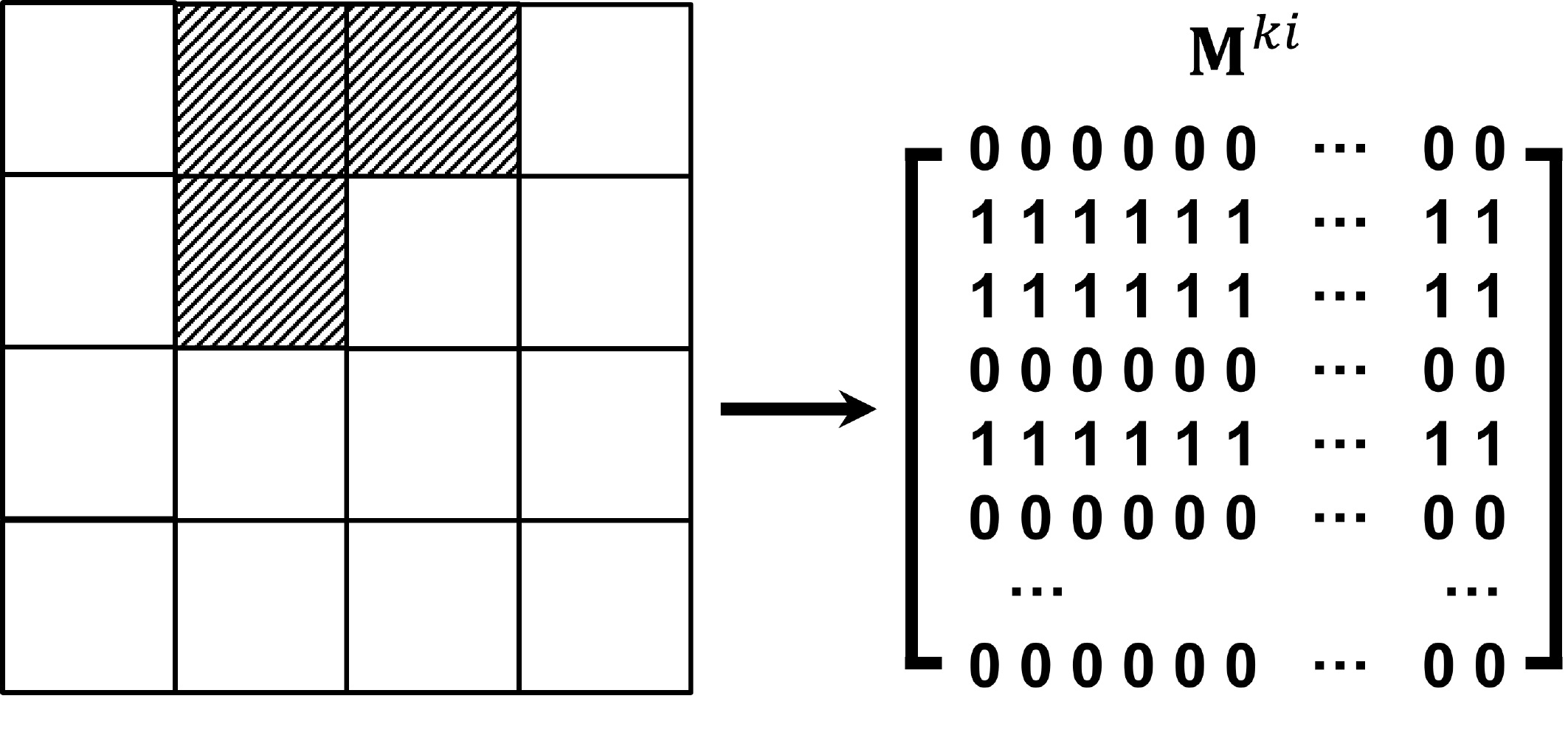}
   \end{tabular}
   \end{center}
   \caption[figure2] 
   { \label{fig:figure2} 
An example of a metal attention matrix for the neighborhood embedding. Each image patch corresponds to each row in the matrix.}
\end{figure}

In order to provide the complementary metal information in the encoder, two types of metal-conscious embedding methods are proposed and the detailed structures are displayed in the lower part of Fig.~\ref{fig:figure1}. Because the cascaded Swin blocks are used, the two embedding methods both contribute to all three Swin blocks. The metal-conscious SE is solely from the metal mask itself. The metal mask is split into different sequences by patches with side length of 4, 8 and 16. Each sequence is then embedded to the corresponding channel size to sum up with the inputs of Swin blocks at different stages, respectively. The NE is conferred on the metal patches by their surrounding patches which are embedded from the input image, and the NE of a metal patch is obtained by an attention layer which has an extra metal attention matrix. After maxpooling of a metal mask with the kernel size $k$, a binary vector $\textbf{m}^{ki}\in\mathbb{R}^{N\times 1}$ can be generated for the N patches in the $i$-th window. Then a corresponding metal attention matrix $\textbf{M}^{ki}$ is obtained as following,
\begin{equation}
\label{metal_att}
\textbf{M}^{ki}=\textbf{m}^{ki}\textbf{I}^{T},
\end{equation}
where $\textbf{I}\in\mathbb{R}^{N\times 1}$ is a vector with ones. An example of such metal attention matrix with the window size of 4 is shown in Fig.~\ref{fig:figure2}, where the shadow patches contain metal. During the process of attention calculation within a window, the metal attention matrix $\textbf{M}^{ki}$ multiplies with the attention matrix $\textbf{A}^{ki}$ from the query and key to get the metal embedding for the Swin ViT block with the patch size of $k$:
\begin{equation}
\label{metal_att}
\textbf{E}_{NE}^{ki}=\textbf{M}^{ki}\textbf{A}^{ki}\textbf{v}^{ki},
\end{equation}
where $\textbf{v}^{ki}$ is the value term. In this case, the rows of $\textbf{E}_{NE}^{ki}$ are zero vectors for the metal-free patches but the embeddings for metal patches are the combinations of their neighbors. The networks are trained for maximum 100 epochs and the training stops when the validation set has the smallest MAE in the metal region.

\subsection{Data generation}
The simulated CBCT projections as well as projections of randomly distributed metallic implants are used for model training. In total, 50 leg volumes are selected from the SICAS medical image repository \cite{kistler2013virtual}. Each volume is forward projected to 60 projections with incremental angular step of 6 degrees. The source-to-detector distance and the source-to-isocentor distance are 1164 mm and 622 mm, respectively. The detector has the size of 512\,$\times$\,512 pixels and the pixel size is 0.580 mm\,$\times$\,0.580 mm. 2700 projections are used for model training and 300 projections are used for validation. In addition, five extra sets of projections are generated for model test. Data argumentation methods are performed during training, for example, random Gaussian noise injection, metal mask dilation and erosion, horizontal or vertical flip. As a preliminary study on the generalizability to real data, knee projections from cadaver scans are also used for model training and test, with 3600 projections for training and one scan of 400 projections for model test.

\section{Results and discussion}
For comparison, the projections in the test data set are inpainted by different methods, which are interpolation, the MPN with normal and gated convolutions, the hybrid ViT-CNN network, and our proposed hybrid Swin-CNN network with metal-conscious embedding. In this case, both the projections and the reconstructed volumes are used for evaluation, where MAE and the peak signal-to-noise ratio (PSNR) are used as evaluation metrics. For better evaluation of the inpainting task, the projections are rescaled to the original intensity window and only metal regions account for the MAE calculation. The quantitative results of different methods are listed in Tab.~\ref{tab1}. As it can be seen, the gated convolutions can improve the performance of MPN, which shows our results are comparable to the work by \cite{ agrawal2021metal}. Moreover, the hybrid Swin-CNN network without metal embedding shows similar performance compared to the MPN with gated convolutions. 
\begin{table}[hbt]
\caption{\label{tab1}Results of different methods}
\centering
\begin{small}
\begin{tabular}{c|cc|cc}
\hline
&\multicolumn{2}{c|}{Projection}&\multicolumn{2}{c}{Reconstruction}\\
\hline
Method & MAE &  PSNR & MAE (HU) &  PSNR\\
\hline
Interpolation & 0.135 &37.204 & 25.403 &35.501\\
\hline
MPN  & 0.111 & 39.606 & 25.089 & 36.394\\
\hline
MPN(gated)  & 0.096 & 40.685 & 24.054 & 37.466\\
\hline
ViT16-CNN  & 0.295 & 31.845 & 47.156 & 30.617\\
\hline
Swin-CNN  &  0.098 & 40.529 & 23.187 & 37.321\\
\hline
\end{tabular}
\end{small}
\end{table}
\begin{table}[hbt]
\caption{\label{tab2}Results comparison among different metal-conscious embedding methods for the Swin-CNN network}
\centering
\begin{small}
\begin{tabular}{c|c|cc|cc}
\hline
\multicolumn{2}{c|}{}&\multicolumn{2}{c|}{Projection}&\multicolumn{2}{c}{Reconstruction}\\
\hline
\multicolumn{2}{c|}{Embedding} & MAE & PSNR & MAE (HU) & PSNR\\
\hline
\multicolumn{2}{c|}{SE}  & 0.088 & 41.323 & 19.081 & 38.227\\
\hline
\multirow{4}{*}{NE} & W4 & 0.092 & 40.985 & 20.201 & 38.321\\
& W8 & 0.087 & 41.381 & 18.625 & 38.780\\
& W16 & \textbf{0.079} & \textbf{42.346} & \textbf{18.021} & \textbf{38.921}\\
& W32 & 0.088 & 41.237 & 20.910 & 38.333 \\
\hline
\end{tabular}
\end{small}
\end{table}

After incorporating the metal-conscious embedding blocks together with the hybrid Swin-CNN network, significant improvements can be observed in Tab.~\ref{tab2}. The MAE in projection domain decreases to 0.088 with the help of the SE. At the same time, the average PSNR increases to 41.891. In the volume domain, the MAE decreases to 17.481 HU and the mean PSNR increases to 45.205. In the case of the NE, the results under different window size are compared. By setting the window size to 16, the model achieves the lowest MAE of 0.079 and the highest mean PSNR of 42.346 in the projection domain. The MAE and PSNR in the reconstruction domain are 16.529 HU and 45.885, respectively. As listed in Tab.~\ref{tab3}, the hybrid Swin-CNN network with metal-conscious embedding also has better performance for the cadaver data set over other methods.
\begin{table}[hbt]
\caption{\label{tab3}Results comparison in cadaver data set}
\centering
\begin{small}
\begin{tabular}{c|cc|cc}
\hline
&\multicolumn{2}{c|}{Projection}&\multicolumn{2}{c}{Reconstruction}\\
\hline
Method & MAE &  PSNR & MAE (HU) &  PSNR\\
\hline
Interpolation & 0.099 &42.082 & 20.759 & 37.426\\
\hline
MPN  & 0.085 & 44.196 & 15.784 & 40.633\\
\hline
MPN(gated)  & 0.080 & 44.694 & 15.380 & 41.037\\
\hline
ViT16-CNN  & 0.254 & 35.605 & 31.243 & 34.539\\
\hline
Swin-CNN  &  0.080 & 44.471 & 14.615 & 41.381\\
\hline
Swin-CNN(SE)  & 0.074 & 45.153 & 13.928 & 41.721\\
\hline
Swin-CNN(NE W16)  &  \textbf{0.069} & \textbf{45.471} & \textbf{13.755} & \textbf{41.841}\\
\hline
\end{tabular}
\end{small}
\end{table}

\begin{figure*}[htb]
   \begin{center}
   \begin{tabular}{c} 
   \includegraphics[width=\textwidth]{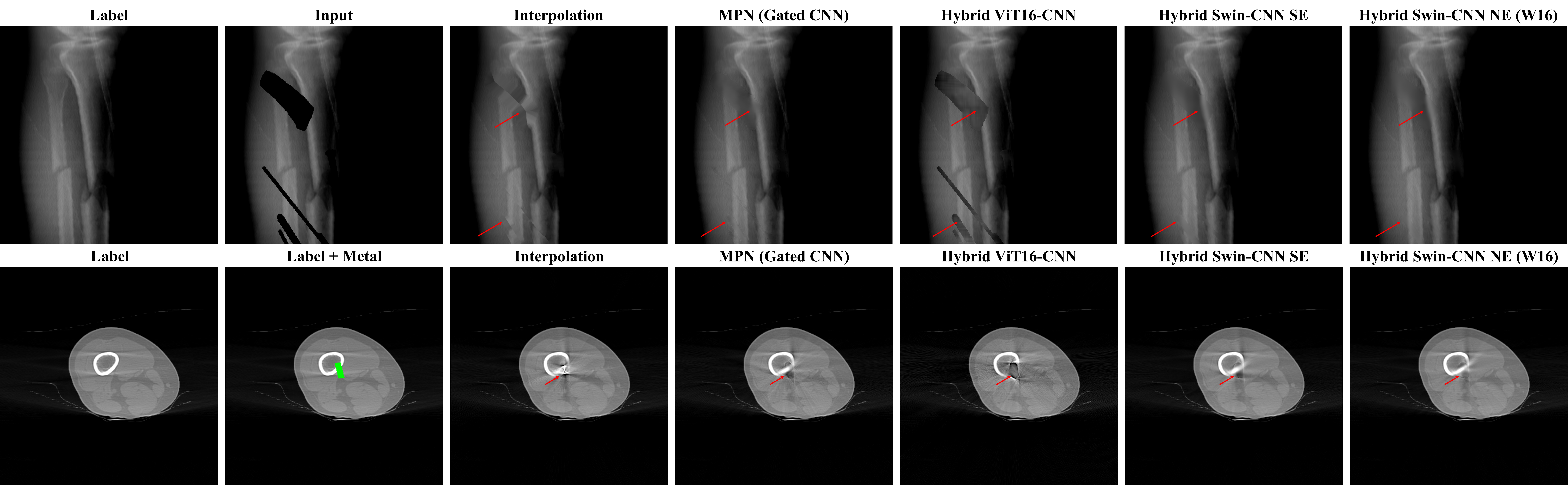}
   \end{tabular}
   \end{center}
   \caption[figure3] 
   { \label{fig:figure3} 
The projection inpainting results (upper, intensity window [2.5, 5]) and the reconstruction results (lower, intensity window [-1000, 1000] HU) by different methods. The streak artifacts in the reference image is caused by the cone angle and the green region indicates the metal area.}
\end{figure*}

Fig.~\ref{fig:figure3} illustrates the inpainting results of one exemplary case. The projection images and reconstructed volumes are displayed in the top and bottom rows, respectively. As it can be seen, the input projection is masked out by both thin and wide metals. The MPN with gated convolutions provides better recovery than interpolation. However, the inpainted areas are slightly blurred and the contours of bones are unclear. Those areas are pointed out by the red arrows.
The hybrid network with the ViT as the encoder leads to the worst result, where the intensity and semantic textures inside the metal regions are clearly wrong. On the other hand, the predictions given by the hybrid Swin-CNN networks with the metal-conscious embedding are able to recover the bone contours well. As pointed out by the red arrow below, the NE with the window size of 16 helps the network achieve more consistent inpainting with the background compared to the result given by the network with SE. Consequently, the hybrid Swin-CNN network with the NE also brings the least artifacts in reconstructions as shown in the lower part of Fig.~\ref{fig:figure3}.

One can conclude that the MPN approach has poor performance when dealing with large metals. This is due to the fact that MPN has relatively small receptive field because of the kernel-based convolution. In addition, the hybrid network with ViT as the encoder fails to accurately recover the missing regions because of the large patch size of 16$\,\times\,$16. In fact, using the large patch size leads to coarse inpainting results. With much smaller patch split, Swin-CNN with the Swin transformer as the encoder is more suitable for the regression task and it can also benefit from a relatively larger receptive field compared with the MPN. After enhanced with the extra information of metals in the encoder, the baseline network has better performance with the help of SE or NE. The SE only gives an initial estimation of the missing areas, and the training process optimizes its estimation ability. Therefore, it has inferiority compared with the NE, which confers the embedding in a more reliable way. In the scenario of the NE, the choice of window size matters. When the window size is too small, not enough surrounding patches are taken into consideration especially in the case of large metals. When the window size is too large, irrelevant patches will account for the embedding of the unknown region, which is also unfavorable. According to our experiment, the hybrid Swin-CNN network has the best performance when the window size is set to 16.

\section{Conclusion}
A hybrid Swin-CNN network with metal-conscious embedding has been proposed for CBCT projection inpainting. The baseline network has comparable performance as the reference network MPN with gated convolutions. By means of the SE and the NE, the performance of our baseline method is further improved. By getting benefit from the NE with the optimal window size, the best model reduces the MAE significantly compared to the reference model MPN with gated convolutions in both simulation and cadaver data, showing its efficacy for CBCT projection inpainting task.

\section{Compliance with ethical standards}
\label{sec:ethics}
This research study was conducted retrospectively using human subject data made available in open access by the SICAS medical image repository \cite{kistler2013virtual}. Ethical approval was not required as confirmed by the license attached with the open access data. This study was performed in line with the principles of the Declaration of Helsinki. Approval was granted by the Ethics Committee of the University Hospital of Erlangen, Germany.


\section{Acknowledgments}
\label{sec:acknowledgments}
This work was supported by academic-industrial collaboration with Siemens Healthineers, XP Division. The presented method is not commercially available and its future availability is not guaranteed.




\bibliographystyle{IEEEbib}
\bibliography{strings,refs}

\end{document}